\documentclass{article} % For LaTeX2e
\usepackage[final]{colm2026_conference}
\usepackage{microtype}
\usepackage{hyperref}
\usepackage{url}
\usepackage{booktabs}
\usepackage{graphicx}
\usepackage{wrapfig}
\usepackage{subcaption}
\usepackage{pgfplots}
\usepackage{tikz}
\usepackage{tabularx}
\usepackage{amsmath}

\usepackage{lineno}

\definecolor{darkblue}{rgb}{0, 0, 0.5}
\definecolor{excerptblue}{HTML}{FBFCFE}
\definecolor{excerptborder}{HTML}{E7EAFA}
\hypersetup{colorlinks=true, citecolor=darkblue, linkcolor=darkblue, urlcolor=darkblue}

\definecolor{groupB}{HTML}{ECEBE5}
\definecolor{grpProg}{HTML}{E7EAFA}      % cobalt tint     -- Programming
\definecolor{grpResearch}{HTML}{E8EEEA}  % deep-sage tint  -- Research & reasoning
\definecolor{grpEnt}{HTML}{F4E8DF}       % sienna tint     -- Enterprise workflows
\definecolor{grpGui}{HTML}{EDE6EF}       % plum tint       -- GUI
\definecolor{grpFC}{HTML}{F3E5E5}        % wine tint       -- Function calling
\definecolor{taskorange}{HTML}{A6531A}

\title{Predicting Task Difficulty Without Rollouts}

\author{Stefan Krsteski \& Charlotte Meyer \\
Andromede AI, EPFL \\
{\{stefan, charlotte\}@andromede.ai} \quad
{\{stefan.krsteski, charlotte.meyer\}@epfl.ch}
}

\newcommand{\taskid}[1]{\textcolor{taskorange}{\emph{#1}}}

\begin{document}

\ifcolmsubmission
\linenumbers
\fi

\maketitle

\begin{abstract}
% Approx how i want it:
% [What does difficulty mean, and what does without rollouts mean in one sweet sentence]
% [Goal: allow for agent evaluation, curriculum design, ... must be aligned with the rest of the paper]
% [Prob: longer-horizons - main motive sentence short 
% [prior work]
% [what we do, compressed]
% [some spoilers/hooks insights but very shallow - can be presented directly takeaways]

% Task difficulty is central to agent evaluation, curriculum design, and behavioral analysis, but it is usually measured by running agents and observing success. As agents solve longer-horizon tasks, this post-hoc measurement becomes costly, motivating predictors that estimate difficulty from task descriptions and metadata before rollouts. We study this problem using 415{,}470 agent-task outcomes across 17 benchmarks, fitting IRT task difficulties and predicting within-benchmark standardized difficulty from structural features, text embeddings, entropy summaries over bounded reasoning traces, and cross-scorer disagreement. Overall, we find that task difficulty is predictable before rollouts, but robust cross-benchmark prediction remains unsolved, making difficulty residuals a useful diagnostic target for feasibility, familiarity, and evaluator effects.

% Task difficulty dictates an agent's likelihood of success, and estimating it without rollouts means forecasting this directly from a task description before executing costly simulations in stateful environments. 

A fundamental challenge in evaluating and training autonomous agents is measuring the intrinsic difficulty of the tasks they attempt. Estimating this quantity can be useful for environment designers creating synthetic data or benchmarks, as well as for constructing training curricula, thereby offering a way to reduce compute costs. Such an estimation becomes increasingly important as agents (particularly LLM-based) move into longer-horizon domains, where empirical trial-and-error becomes a severe computational bottleneck. However existing work is largely confined to static tasks and relies on evaluation metrics that, as we show, can give an incomplete picture of predictive performance. In this paper we study \textit{ex ante} difficulty prediction across 17 agentic benchmarks spanning coding, mathematics, machine learning, web navigation, function calling, and other domains. We find that AUC can mask poor difficulty estimates, identify token-level entropy as a useful predictive signal, and demonstrate how residuals between expected and observed difficulty can expose environment flaws such as contamination and infeasibility.

\end{abstract}

\section{Introduction}
\label{sec:intro}

As LLM agents rapidly improve, the performance frontier is pushed toward domains that are increasingly hard to design, validate, and simulate. This shift creates a financial and logistical bottleneck for environment designers. Evaluating frontier systems on complex, long-horizon tasks can require hours of simulated interaction and many model generations per attempt \citep{kwa2026measuring}. Consequently, the field faces a paradox: progress requires carefully calibrated environments, but verifying their difficulty empirically is increasingly prohibitive. This paper argues for a shift toward \textit{ex ante} difficulty prediction, with the goal of estimating whether a task is likely to be hard before spending large amounts of compute on evaluation or training.

Recent work has used a model's uncertainty to detect data contamination in text-based benchmarks like question-answering \citep{li2309estimating,xu2024benchmark,dong2024generalization}. Yet contamination can be viewed as an instance of a more fundamental variable, namely task difficulty. If a model has effectively memorized an item, the task becomes easier for reasons unrelated to the intended capability. In interactive environments, difficulty can affect both success rates and the strategies agents use. In human settings, high task complexity can push people to alter their strategies, such as resorting to dishonesty \citep{mazar2008dishonesty}. Similarly, when language-model agents operate near the edge of their competence, they may exploit shortcuts \citep{skalse2022defining,helff2026llms}. Additionally, difficulty can reveal when an evaluation task is misaligned with its intended purpose. A task may register as difficult because it exceeds an agent's capabilities, or simply because the environment precludes a successful solution. Conversely, a task may appear easy because it is familiar or prone to shortcuts. We therefore posit that difficulty serves as a common variable connecting task design and agent behavior.

If difficulty could be estimated before rollouts, environment designers could construct calibrated evaluations without first spending the cost of full empirical validation. Training pipelines could also build curricula of progressively harder tasks \citep{bengio2009curriculum}, while evaluations could better contextualize success and failure \citep{kwa2026measuring}.

Despite this motivation, pre-rollout difficulty prediction remains underdeveloped. To the best of our knowledge, only one prior paper directly studies this setting \citep{ge2026agent}, specifically in coding benchmarks. This leaves two open questions: how should such estimates be evaluated, and do they transfer across benchmark families? We show that response AUC can be misleading, while task-level rank metrics give a more cautious picture. In this work, we broaden the setting beyond coding tasks to seven additional benchmark types and take a diagnostic view of difficulty prediction. Our core contributions are:

% TODO i think these can be better
\begin{itemize}
    \item \textbf{Evaluation of difficulty prediction}, which shows that response AUC can remain high even when task difficulties are poorly estimated. Under rank correlation, entropy is the strongest single feature family, while the combined feature set performs best overall, reaching Spearman \(\rho=0.399\) under K-fold evaluation (\(p<0.001\) vs. baseline) and \(\rho=0.225\) out of distribution (\S\ref{sec:evaluation}, \S\ref{sec:results}).

    \item \textbf{Residual diagnostics for hidden task factors}, showing how gaps between predicted and observed difficulty can guide follow-up analysis. Positive residuals point to unexpectedly hard tasks, such as feasibility or evaluator failures, while negative residuals point to unexpectedly easy tasks, such as familiarity (\S\ref{sec:results-contamination}).

    \item \textbf{An open multi-benchmark corpus} with 415{,}470 agent-task outcomes across 17 benchmarks, including task metadata, verifier summaries, textual descriptions, and IRT-estimated difficulties for reuse beyond this paper in behavioral analysis, environment design, and training (\S\ref{sec:methods-data}).
\end{itemize}

\section{Related work}
\label{sec:related}

Taking the rollout as the point of reference, we distinguish three sets of methods by when the difficulty signal is observed: after, during, and before execution.

\paragraph{After rollout.}
Agentic benchmarks often use empirical success rates, rewards, or other trajectory-level quantities such as number of actions, steps, and elapsed time to estimate difficulty, with human completion time serving as a gold standard \citep{kwa2026measuring,liu2026bridge,ho2025rosetta}. These methods make difficulty explicit, but their signal is observed only after agents or humans interact with the task. Our work instead studies whether difficulty can be predicted from information available before running agents.

\paragraph{Adaptive methods.}
Curriculum learning relies on difficulty to order training examples or environments for greater sample efficiency \citep{bengio2009curriculum,vygotsky1978zpd,portelas2020automatic}. Many online variants define difficulty relative to the current learner. Some methods, such as self-paced and automated curriculum learning, use loss, prediction error, or temporal-difference (TD) error as the difficulty signal \citep{kumar2010self,graves2017automated,jiang2021prioritized}. Others, such as teacher-student curricula and unsupervised environment design, use reward, regret or success to select tasks that are neither too easy nor too hard \citep{florensa2018automatic,dennis2020emergent,parker2022evolving,beukman2024refining,tio2023training,mahrooghi2026goldilocks,matiisen2019teacher,bae2026online}. Nonetheless, these methods require executing tasks, often inside an additional teacher-student loop, which can further compound the rollout overhead we aim to circumvent.

\paragraph{Before rollout.}
More distant work on static benchmarks motivates using uncertainty signals from language models, including perplexity and entropy \citep{gonen2023demystifying,li2309estimating,xu2024benchmark,dong2024generalization,spiesberger2026soft,zhu2026edis,ostmeier2026attention,catala2026stepwise,feng2026generating,truong2025reliable}. These settings are not interactive and do not cover agentic benchmarks. To the best of our knowledge, Agent Psychometrics is the only prior work that directly studies \textit{ex ante} difficulty prediction for agentic tasks, focusing on coding benchmarks \citep{ge2026agent}. We test broader generalization by evaluating predictors across seven additional benchmark families. We also evaluate a broader feature set, including token-level entropy, cross-scorer disagreement, structural features, and embeddings.

% We must properly differentiate ourselves from Agent Psychometrics - we are an extension - we use more features, more experiments, show their metrics don't work (What about their baseline? Just not overly aggressive)
% Agent psychometrics do only coding environments, whereas we expand the scope Agent psychometrics use only embeddings and 1 model, where we expand the features (using logprobs, and the amount of models) Agent psychometrics do 1PL, where we incorporate BRIDGE’s 2PL approach

\section{Methods}
\label{sec:method}

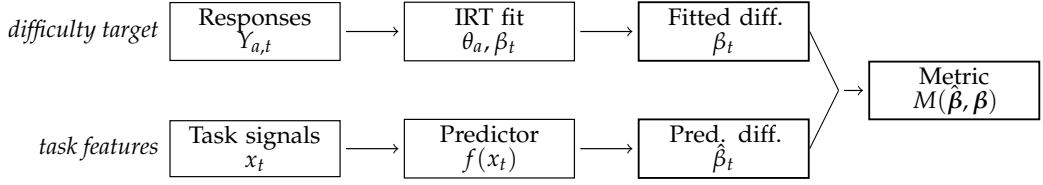
\begin{figure}[t]
  \centering
  \begin{tikzpicture}[
    box/.style={draw=black, fill=white, line width=0.45pt,
      minimum width=2.25cm, minimum height=0.72cm, text width=1.95cm,
      inner sep=2pt, align=center},
    output/.style={box, line width=0.7pt},
    eval/.style={box, line width=0.7pt},
    arrow/.style={->, line width=0.45pt, shorten >=2pt, shorten <=2pt},
    font=\small
  ]
    \node[anchor=east, font=\small\itshape] at (-1.15, 0)
      {difficulty target};
    \node[anchor=east, font=\small\itshape] at (-1.15, -1.55)
      {task features};

    \node[box] (responses) at (0, 0) {Responses\\[-1pt] $Y_{a,t}$};
    \node[box] (irt) at (3.1, 0) {IRT fit\\[-1pt] $\theta_a,\beta_t$};
    \node[output] (target) at (6.2, 0) {Fitted diff.\\[-1pt] $\beta_t$};

    \node[box] (features) at (0, -1.55) {Task signals\\[-1pt] $x_t$};
    \node[box] (predictor) at (3.1, -1.55) {Predictor\\[-1pt] $f(x_t)$};
    \node[output] (predicted) at (6.2, -1.55) {Pred. diff.\\[-1pt] $\hat{\beta}_t$};

    \node[eval] (metrics) at (9.25, -0.78) {Metric\\[-1pt] $M(\hat{\boldsymbol{\beta}},\boldsymbol{\beta})$};

    \draw[arrow] (responses) -- (irt);
    \draw[arrow] (irt) -- (target);
    \draw[arrow] (features) -- (predictor);
    \draw[arrow] (predictor) -- (predicted);
    \coordinate (merge) at (7.72, -0.78);
    \draw (target.east) -- (merge);
    \draw (predicted.east) -- (merge);
    \draw[arrow] (merge) -- (metrics.west);
  \end{tikzpicture}
    \caption{\textbf{Overview of the difficulty prediction pipeline.} Upper branch: binary agent success and failure outcomes are used to fit IRT and estimate each task's difficulty, \(\beta_t\). Lower branch: task features are used to predict difficulty, \(\hat{\beta}_t\), and evaluation compares the predicted and IRT-estimated difficulty vectors.}  \label{fig:target-pipeline}
\end{figure}

We decompose the prediction pipeline into two steps: constructing a difficulty target from observed agent-task outcomes, then training a predictor to estimate this target from task descriptions and other pre-rollout features (Figure~\ref{fig:target-pipeline}). 

\subsection{Dataset}
\label{sec:methods-data}

\begin{wraptable}[15]{r}{0.40\textwidth}
  \centering
  \small
  \begin{tabularx}{\linewidth}{@{}X r@{}}
    \toprule
    \textbf{Statistic} &
    \textbf{Value} \\
    \midrule
    \# Benchmarks & 17 \\
    \# Benchmark types & 8 \\
    \# Tasks & 5,230 \\
    \# Models & 216 \\
    \# Scaffolds & 90 \\
    \# Agents & 497 \\
    Task dates & 2010--2026 \\
    Agent dates & 2023--2026 \\
    Avg. trials / agent-task & 1.2 \\
    Mean success rate & 0.539 \\
    \bottomrule
  \end{tabularx}
  \caption{Summary of the dataset.}
  \label{tab:data-summary}
\end{wraptable}

Our dataset~\citep{krsteski2026messier}\footnote{This work uses a preliminary version of the corpus.} consists of agent-task outcomes from 17 agentic benchmarks, listed in Appendix~\ref{app:benchmark-list}. Each instance records one trial of one agent configuration on one task, labeled by success under the benchmark's scoring rules. We use \emph{agent} to mean a model (LLM) together with the scaffold through which it acts.

Table~\ref{tab:data-summary} summarizes the main analysis subset used throughout the main-text results. It contains 5{,}230 tasks from 17 benchmarks, evaluated by 497 agent configurations built from 216 models and 90 scaffolds. For each task, the task, environment, and action-space fields are stored as string descriptions derived from the benchmark, either directly or via an LLM when missing. These benchmarks span coding and software engineering, function calling, web navigation, computer use, terminal tasks, math, ML, cybersecurity, and general-purpose task suites. Task dates denote the task release or source date when available and range from 2010 to 2026. Agent dates denote the model-weight release date when available, otherwise the first public availability date, and range from 2023 to 2026. Each trial is treated as an independent Bernoulli observation under the benchmark scoring rules, with 1.2 trials per agent-task pair on average.

\subsection{Difficulty target}
\label{sec:methods-target}

IRT is a psychometric framework that models the relationship between an individual's latent trait (e.g., knowledge, ability, anxiety) and their responses to specific test items \citep{embretson2025item}. In our case, the ``individual'' is the agent and the ``test items'' are the tasks. Let \(\mathcal{A}={a_1,\ldots,a_N}\) denote the set of ($N$) agent configurations and \(\mathcal{T}={t_1,\ldots,t_M}\) the set of ($M$) tasks. The data form a partially observed binary response matrix
\[
Y_{a,t} =
\begin{array}{c@{\;}c}
&
\begin{array}{ccccc}
t_1 & t_2 & t_3 & \cdots & t_M
\end{array}
\\[0em]
\begin{array}{c}
a_1 \\ a_2 \\ \vdots \\ a_N
\end{array}
&
\left[
\begin{array}{ccccc}
1 & 0 & \mathrm{x} & \cdots & \mathrm{x} \\
0 & 1 & 0          & \cdots & \mathrm{x} \\
\vdots & \vdots & \vdots & \ddots & \vdots \\
\mathrm{x} & \mathrm{x} & 1 & \cdots & 1
\end{array}
\right]
\end{array}
\]

where $1$ denotes success, $0$ denotes failure, $\mathrm{x}$ denotes a missing evaluation, and \(\Omega\) is the set of observed agent-task pairs. Each row is a unique agent configuration, such as a base model with a scaffold, and each column is a unique task. The goal is to model the probability of success by learning one scalar $\theta_a$ for each agent and one scalar $\beta_t$ for each task. In the one-parameter logistic IRT model (1PL; Rasch model \citep{rasch1960probabilistic}), this is written as
\begin{equation}
P(Y_{a,t}=1 \mid \theta_a,\beta_t)=\sigma(\theta_a-\beta_t),
\end{equation}

where $\theta_a$ represents agent capability, $\beta_t$ represents task difficulty and \(\sigma(z)=1/(1+\exp(-z))\) is the sigmoid function. This form gives the desired inductive bias because success depends on the relative position of capability and difficulty on a shared scale. If $\theta_a=\beta_t$, capability and difficulty are balanced; if $\theta_a>\beta_t$, capability exceeds difficulty; and if $\theta_a<\beta_t$, difficulty exceeds capability. The resulting \(\beta_t\) is therefore a scalar summary of apparent task difficulty as inferred from observed outcomes, and may absorb task-side covariates such as feasibility, strictness, familiarity, and noise, just as \(\theta_a\) may absorb agent-side covariates such as scaffold effectiveness, tool affordances, and noise. With an interpretable predictor and an appropriate setup, this lets us ask which features account for different parts of the fitted difficulty.

% More expressive IRT variants, such as two-parameter logistic models \cite{lord2012applications}, could separate additional factors, but we use 1PL for interpretability and leave richer parameterizations to future work.

The model is trained on observed pairs $(a,t)\in\Omega$ by optimizing the evidence lower bound (ELBO) via stochastic variational inference (SVI) \citep{hoffman2013stochastic} and the reparameterization trick \citep{kingma2013auto}. We place zero-centered Gaussian priors over all parameters and approximate the posterior distribution over these parameters with independent Gaussian variational distributions:
\[
  q(\theta_a)=\mathcal{N}(\mu_{\theta_a},\sigma_{\theta_a}^2), \qquad q(\beta_t)=\mathcal{N}(\mu_{\beta_t},\sigma_{\beta_t}^2).
\]
Because the model depends strictly on the difference $\theta_a - \beta_t$, the parameters can shift arbitrarily without changing the probabilities. To resolve this, we anchor the scale by centering the abilities $\theta_a$ at zero. We then use the resulting task difficulty estimates ($\beta$) as the target for downstream prediction. The fitted scores achieve a response AUC of $0.937$ on observed pairs, indicating that the 1PL model captures much of the observed response structure. Finally, as difficulties vary across different benchmarks, we compute a standardized version of these values within each benchmark to obtain $z(\beta_t)$. More details are provided in Appendix~\ref{app:irt-fitting-details}.

\subsection{Task features}
\label{sec:methods-features}

The target is the IRT-estimated difficulty \(\beta_t\), which we standardize within-benchmark to \(z(\beta_t)\) for our main experiments, as described above. The practical constraint is that every feature must be less expensive to compute than running rollouts on the task. Let \(C_t\) denote the full task-side context for task \(t\), consisting of the environment description, action-space description, and task description. Feature families use \(C_t\) either directly or through a reasoning trace \(R_t\) generated from \(C_t\) by Claude Sonnet 4.6 with a 4k-token cap.

We use five feature sets throughout the experiments. The \emph{random} control samples \(x_t \sim \mathcal{N}(0,1)\) independently from the task. The \emph{baseline} setting uses only the context length, \(x_t^{\text{len}}=\log(1+|C_t|)\), to control for the amount of task text. The \emph{embedding} family maps the reasoning trace to \(e_t=\phi(R_t)\), where \(\phi\) is \texttt{bge-base-en-v1.5}. We reduce this representation with fold-aware PCA to avoid overfitting.

The \emph{entropy} setting uses an open-weight scorer model. Given \(R_t=(r_1,\ldots,r_L)\), the scorer defines \(p_\ell(v)=p(v\mid C_t,r_{<\ell})\) at each position \(\ell\), where \(v\) is a token from the model's vocabulary. We compute Shannon entropy \citep{shannon1948mathematical} as
\[
  H_\ell=-\sum_v p_\ell(v)\log p_\ell(v),
\]
using the renormalized top-100 next-token distribution for faster computation. The primary entropy configuration aggregates \((H_1,\ldots,H_L)\) from Qwen3-8B-Base over the reasoning trace \(R_t\). The \emph{full} setting concatenates entropy summaries from a five-model scorer panel, cross-scorer disagreement features, benchmark metadata, structural properties of \(C_t\), and the embedding features defined above. Appendix~\ref{app:entropy-scorers} gives the scorer panel.

To see that these features satisfy the cost constraint from above, consider a 100-task benchmark. Establishing empirical difficulty at a density of 10 outcomes per task requires 1{,}000 full executions. Assuming a representative 100{,}000-token context per execution \citep{bai2026ai}, this process demands approximately \(1.0 \times 10^8\) input tokens. In contrast, estimating difficulty \textit{ex ante} using the features proposed here (accounting for the reasoning trace) consumes roughly \(4.0 \times 10^6\) tokens for the same 100 tasks. Under these assumptions, the feature pass uses \(25\times\) fewer tokens than empirical difficulty estimation, and the efficiency grows with the number of outcomes required per task.

\section{Evaluation setup}
\label{sec:evaluation}

\subsection{Prediction metrics}
\label{sec:methods-evaluation}

The preceding sections define a target task difficulty \(\beta_t\), obtained from observed agent-task outcomes, and a predicted difficulty \(\hat{\beta}_t=f(x_t)\), obtained from task features available before running rollouts. The evaluation asks how well \(\hat{\beta}_t\) recovers \(\beta_t\).

We report Spearman's \(\rho\) between predicted and IRT-estimated task difficulty. We use \(\rho_{\beta}\) for raw fitted difficulty and \(\rho_{z(\beta)}\) for within-benchmark standardized difficulty. We also report within-benchmark pairwise accuracy (\(\mathrm{pAcc}\)) as a complementary ordering metric. Let \(q_t\) denote the target being evaluated, either \(\beta_t\) or \(z(\beta_t)\), and let \(\hat{q}_t\) denote its prediction. Then \(\mathrm{pAcc}\) is the fraction of within-benchmark task pairs for which the predicted ordering matches the target ordering:
\begin{equation}
\mathrm{pAcc}
=
\frac{1}{|\mathcal{P}|}
\sum_{(i,j)\in\mathcal{P}}
\mathbb{1}\left[
(\hat{q}_i-\hat{q}_j)(q_i-q_j)>0
\right],
\end{equation}

where \(\mathcal{P}\) is the set of task pairs within the same benchmark. K-fold evaluation (KF) measures in-distribution performance, where tasks from the target distribution are available during training. Leave-one-benchmark-out (LOBO) measures out-of-distribution performance, where the target benchmark is unseen during training.

\subsection{AUC can be misleading}
\label{sec:evaluation-auc-degeneracy}

Prior work \citep{ge2026agent} evaluates difficulty predictors through AUC. Using its probabilistic interpretation, AUC measures the probability that a randomly chosen successful attempt receives a higher score than a randomly chosen unsuccessful attempt. For a successful attempt by agent \(a\) on task \(t\), and an unsuccessful attempt by agent \(a'\) on task \(t'\), the IRT score gives
\begin{equation}
\label{eq:auc-decomposition}
\begin{aligned}
\mathrm{AUC}
&= P(\theta_a-\hat{\beta}_t > \theta_{a'}-\hat{\beta}_{t'}) \\
&= P(\theta_a-\theta_{a'} > \hat{\beta}_t-\hat{\beta}_{t'}).
\end{aligned}
\end{equation}

This expression shows that AUC here is affected by both the difference in agent abilities and the difference in predicted task difficulties. Consider a degenerate case where the predictor simply assigns the same prediction to all tasks. This makes \(\hat{\beta}_t=\hat{\beta}_{t'}\) for every pair of sampled attempts, meaning
\begin{equation}
\label{eq:auc-constant-difficulty}
\mathrm{AUC}
=
P(\theta_a>\theta_{a'}).
\end{equation}

Thus, AUC can remain high even when the predictor contains no task-specific difficulty information. We demonstrate this failure mode empirically on the observed response matrix by comparing fixed difficulty assignments against the fitted IRT target and observed responses, shown in Table~\ref{tab:auc-degeneracy}.
\begin{table}[t]
  \centering
  \setlength{\tabcolsep}{10pt}
  \begin{tabular}{@{}l rrrr@{}}
    \toprule
    \textbf{Estimator} &
    \textbf{AUC $\uparrow$} &
    \textbf{$\rho_{\beta} \uparrow$} &
    \textbf{$\rho_{z(\beta)} \uparrow$} &
    \textbf{pAcc $\uparrow$} \\
    \midrule
    Oracle &
    0.937 &
    1.000 &
    1.000 &
    1.000 \\
    Bench. avg. &
    0.806 &
    0.519 &
    -- &
    0.000 \\
    Constant &
    0.715 &
    -- &
    -- &
    0.000 \\
    Shuffled &
    0.610 &
    -0.011 &
    -0.005 &
    0.500 \\
    \bottomrule
  \end{tabular}
  \caption{\textbf{Response AUC remains high for poor difficulty estimates.} Benchmark avg. assigns each task its benchmark-average. \(\rho_{\beta}\) and \(\rho_{z(\beta)}\) compare against raw and within-benchmark-standardized \(\beta\). pAcc is within-benchmark pairwise accuracy. Dashes indicate undefined. Note that this is illustrative, all later predictions target \(z(\beta)\) instead of \(\beta\).}
  \label{tab:auc-degeneracy}
\end{table}
The oracle row uses the IRT-estimated $\beta$, establishing an upper bound of 0.937 AUC for this response matrix. Predicting a single constant difficulty for every task still yields an AUC of 0.715, confirming the constant difficulty case in Equation~\ref{eq:auc-constant-difficulty}. This inflation is further exaggerated by the benchmark-average estimator. Simply assigning every task the mean difficulty of its parent benchmark achieves an AUC of 0.806 and a non-trivial rank correlation ($\rho_{\beta}$ of 0.519). This estimator cannot differentiate tasks \textit{within} the benchmark, giving a strict pAcc of 0.000. Because the practical goal is to help benchmark builders rank and select candidate tasks within their specific evaluation settings, recovering this signal is essential. To remove this confound and evaluate task-level difficulty more directly, all subsequent training and prediction target within-benchmark standardized difficulty, $z(\beta)$.

\section{Results}
\label{sec:results}

\subsection{Difficulty is predictable from task features}
\label{sec:results-difficulty-prediction}

\begin{wrapfigure}{r}{0.52\textwidth}
  \centering
  \includegraphics[width=\linewidth]{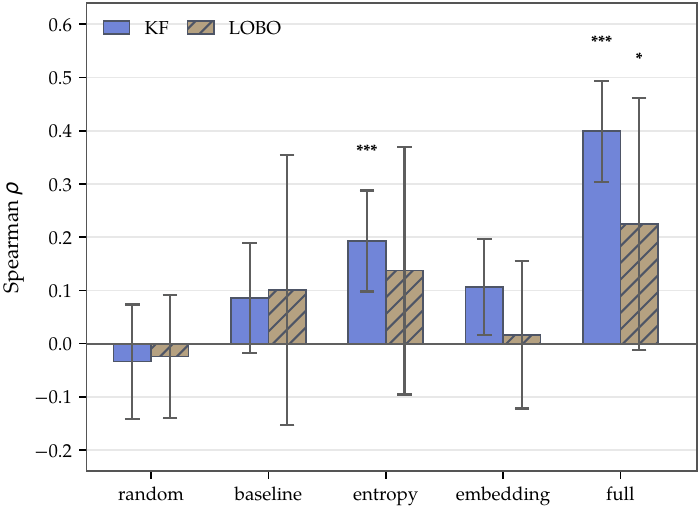}
  \caption{\textbf{Difficulty prediction performance.} The target is \(z(\beta)\). Spearman \(\rho\) is shown under 50-fold KF and LOBO evaluation. Error bars show split standard deviations. With 50 folds, each test fold contains roughly 100 tasks, leading to substantial uncertainty. Stars mark one-sided paired Wilcoxon tests against the baseline at significance levels \(^{*}\alpha=0.05\), \(^{**}\alpha=0.01\), and \(^{***}\alpha=0.001\).}
  \label{fig:difficulty-prediction}
\end{wrapfigure}

Figure~\ref{fig:difficulty-prediction} reports rank recovery across the 17-benchmark corpus using Spearman \(\rho\) and ridge regression (Appendix~\ref{app:estimator-ablation} compares alternative estimators). KF measures in-distribution performance, where similar tasks are available during training. LOBO measures out-of-distribution performance on an unseen target benchmark. The prediction target is \(z(\beta)\).

The random control stays near zero, while the length baseline captures only weak signal with \(\rho=0.086\) under KF and \(\rho=0.101\) under LOBO. The entropy features improve over this baseline, reaching \(\rho=0.193\) under KF and \(\rho=0.137\) under LOBO. Embedding features are weaker in this setup, with \(\rho=0.107\) under KF and \(\rho=0.017\) under LOBO, close to the length baseline. This suggests that these semantic representations capture little additional difficulty signal here, although stronger embedding models may change this conclusion. Combining feature families gives the strongest KF result, \(\rho=0.399\). Under LOBO, the same full model reaches \(\rho=0.225\). Thus, task features encode real difficulty signal, but transfer to unseen benchmarks remains limited. 
Taken together, the results point to a practical near-term setting in which an evaluated task suite already exists (with executed rollouts) and new candidate tasks can be ranked with moderate confidence.

\subsection{Entropy ablations}
\label{sec:results-entropy}

\begin{figure}[t]
  \centering
  \includegraphics[width=\linewidth]{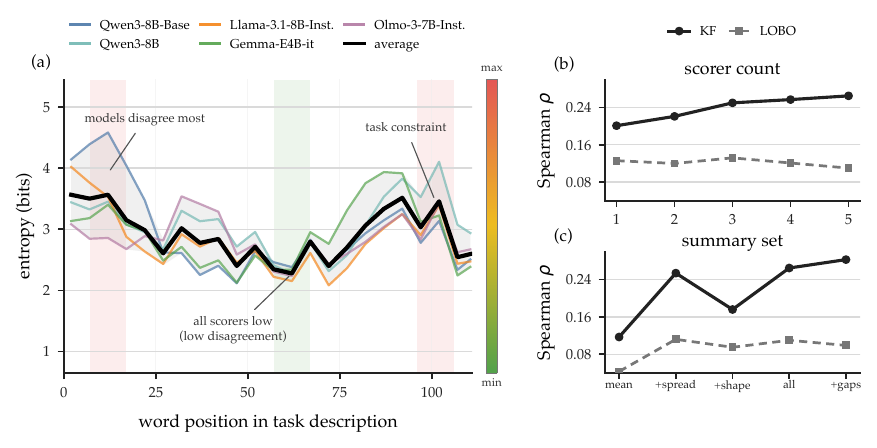}

  \vspace{0.15em}
  {\setlength{\fboxsep}{2pt}
  \fcolorbox{excerptborder}{excerptblue}{%
    \begin{minipage}{0.94\linewidth}
      \scriptsize
      \raggedright
      \textbf{Task (a).}
Consider an axis-aligned $4$ dimensional cube with side length $4$, subdivided into \colorbox{grpFC}{$4^{4}$ unit cubes}. Cubey starts at a random unit cube. A fish is independently placed in a uniformly random unit cube, \colorbox{grpResearch}{possibly the same} as Cubey's starting position. Compute the
      \colorbox{grpEnt}{expected time} for Cubey to catch the fish,
assuming he \colorbox{grpFC}{moves optimally}.
    \end{minipage}%
  }}

  \caption{\textbf{Entropy profiles and feature ablations.} \textbf{(a)} An example task from MathArena, showing the entropy profile over its span. Colored lines are different models, black is their average, and the textual excerpt is given in the blue box. \textbf{(b)} Scorer-count ablation using one fixed order. \textbf{(c)} Shows an ablation of feature derivation from the entropy profile.}
  \label{fig:entropy-ablations}
\end{figure}

As illustrated in Figure~\ref{fig:entropy-ablations}\textbf{(a)}, entropy can be considered a trajectory rather than a static metric. Following the principle that conditioning reduces entropy \citep{cover1999elements}, additional context should reduce expected next-token uncertainty. Consequently, global statistics such as the mean and variance may miss local structure in the entropy profile. For instance, a highly unexpected, difficulty-inducing condition that appears late in the context may have a lower entropy magnitude than the initial tokens. Thus, any aggregate (compression) is susceptible to losing this information.

The critical information may therefore lie in local discrepancies and structural anomalies. Figure~\ref{fig:entropy-ablations}\textbf{(c)} supports this intuition. Beyond the mean, \emph{spread} adds distributional summaries such as standard deviation, quantiles, and top-\(k\) means, \emph{shape} adds trajectory summaries such as slope, first- and last-decile means, and total variation, and \emph{gaps} adds pairwise absolute differences between scorer summaries. These trajectory summaries improve in-distribution performance over mean-only entropy while preserving or improving LOBO performance. This suggests that the shape of the entropy trajectory carries useful information. Future work could model entropy explicitly as a time series, for example with frequency-domain or learned sequence features. With sufficient data, deep learning architectures such as 1D-CNNs or RNNs could also learn such features directly.

At the same time, ensembling models introduces a counterintuitive dynamic. Figure~\ref{fig:entropy-ablations}\textbf{(b)} shows that increasing the number of scorers strengthens the in-distribution performance, while LOBO performance is non-monotonic and does not improve in the same way. One plausible explanation is benchmark-specific overfitting. Averaging across multiple models reduces scorer-specific noise, but it may also reinforce shared sensitivity to benchmark-specific phrasing and formatting, limiting out-of-distribution transfer.

\subsection{Contamination and feasibility}
\label{sec:results-contamination}

One diagnostic use of \textit{ex ante} difficulty prediction is to compare predicted and observed difficulty after rollouts. Let \(z(\beta_t)\) be the within-benchmark standardized difficulty implied by rollouts, and let \(\hat{z}_t\) be the predicted task difficulty from the full ridge model in Section~\ref{sec:results-difficulty-prediction}. Their difference
\begin{equation}
  r_t = z(\beta_t) - \hat{z}_t
\label{eq:difficulty-residual}
\end{equation}

is the unexplained difficulty. If \(r_t>0\), agents found the task harder than the model predicted. If \(r_t<0\), agents found the task easier than the model predicted. The second case is the direction expected under familiarity or contamination, where agents succeed more often than the predicted difficulty would suggest. The first case is the direction expected when some missing factor, such as feasibility, makes the rollout harder than the task description suggests.

To illustrate this idea, we perform a small behavioral case study. An OpenAI audit reports that \(59.4\%\) of 138 frequently failed SWE-bench Verified tasks had material test-design or problem-description issues, and it also reports evidence of contamination for several tasks. Two of these audited examples appear in our dataset. For \taskid{django\_\_django-14725}, flagged for missing prompt information, the residual equation yields \(r_t=1.46-0.05=+1.41\). The task has success rate \(0.090\) and IRT estimated difficulty of \(\beta=2.58\), above the \(90\)th percentile within SWE-bench Verified. For \taskid{django\_\_django-11451}, discussed as a contamination example, the residual equation gives \(r_t=-0.51-0.05=-0.56\). The task has success rate \(0.718\) and IRT-estimated difficulty \(\beta=-2.35\), below the benchmark median. Both residual signs match the intended interpretation. Unexpectedly hard tasks point toward feasibility or evaluator mismatch, while unexpectedly easy tasks point toward familiarity or contamination. 

Because this residual uses empirical difficulty $\beta_t$ from full agent rollouts, it supports detection rather than pure pre-rollout prediction. It remains an open question how far this mechanism can be pushed to automatically audit benchmark integrity and whether such anomalies can eventually be predicted without requiring execution at all.

\section{Conclusion}
\label{sec:conclusion}
We investigate whether the difficulty of agentic tasks can be estimated prior to execution. Our results show both promise and clear limits. Task features, particularly the sequential structure of token-level entropy, capture a meaningful difficulty signal, yet the results still reveal a clear divide between in- and out-of-distribution performance. While in-distribution rank recovery proves reliable, transferring these predictions to unseen benchmarks remains an open problem. We also show that AUC is unreliable for this setting. Because it conflates agent ability with task difficulty, it can remain artificially high even when estimates are degenerate, underscoring the need for rank-based evaluation.

Overall, we believe this work provides a foundation for more proactive environment design. If refined, \textit{ex ante} prediction could serve as a diagnostic filter by using the residual gap between expected and observed difficulty to audit tasks for anomalies, including data contamination or structural infeasibility, before committing to exhaustive rollouts. Future work should focus on improving cross-benchmark generalization, for example by modeling entropy as a time series or by developing more mechanistic predictors of task familiarity and feasibility. As the cost of simulating frontier agents grows, such predictive diagnostics will become increasingly important for both training and evaluation.

\clearpage

% We study whether task difficulty in agentic benchmarks can be estimated before evaluating new target agents. The answer is mixed. Task features recover meaningful difficulty structure, especially under KF, but held-out-benchmark transfer remains substantially weaker. We also show that AUC can be optimistic for this problem because response ranking can be driven by agent ability even when the task-difficulty estimate is degenerate. The practical recommendation is simple: report task-level difficulty metrics, use ex ante predictors to rank and audit tasks before expensive runs, and treat cross-benchmark transfer as unproven until it is validated on the target benchmark.

% Moving forward, promising directions include scaling the scorer-model panel, fitting hierarchical difficulty models that share strength across related agents and benchmarks, and adding state-level task descriptors such as visible state, available choices, and action constraints. Beyond difficulty prediction, these directions point toward cheaper diagnostics for agent evaluation settings where exhaustive rollouts are scarce.

\bibliography{colm2026_conference}
\bibliographystyle{colm2026_conference}

\clearpage

\appendix
\section{Supplementary material}
\label{app:supplementary-results}

\subsection{Benchmark list}
\label{app:benchmark-list}

\begin{table}[h]
  \centering
  \small
  \setlength{\tabcolsep}{4pt}
  \begin{tabularx}{\linewidth}{@{}lX@{}}
    \toprule
    \textbf{Benchmark} & \textbf{Primary domain} \\
    \midrule
    BFCL Live~\citep{patil2025berkeley} & Function calling \\
    BFCL Multi-Turn~\citep{patil2025berkeley} & Multi-turn function calling \\
    CyBench~\citep{zhang2025cybench} & Cybersecurity tasks \\
    GDPval~\citep{patwardhan2025gdpval} & Economically valuable real-world tasks \\
    GSO~\citep{shetty2026gso} & Software optimization \\
    HCAST~\citep{rein2025hcast} & Human-calibrated autonomy tasks \\
    HLE~\citep{phan2025humanity} & Broad knowledge and reasoning \\
    LiveCodeBench~\citep{jain2025livecodebench} & Code generation \\
    MathArena~\citep{balunovic2026matharena} & Competition mathematics \\
    MLE-bench~\citep{chan2025mle} & Machine-learning engineering \\
    Online-Mind2Web~\citep{xue2025illusion,deng2023mind2web} & Web navigation \\
    RE-Bench~\citep{wijk2024re} & Research-engineering tasks \\
    SWAA~\citep{kwa2026measuring} & Short-horizon autonomous-agent tasks \\
    SWE-bench Pro~\citep{deng2025swe} & Software engineering \\
    SWE-bench Verified~\citep{jimenez2024swebench} & Software engineering \\
    Terminal-Bench~\citep{merrill2026terminal} & Terminal tasks \\
    TheAgentCompany~\citep{xu2026theagentcompany} & Office-style multi-tool tasks \\
    \bottomrule
  \end{tabularx}
  \caption{Benchmarks and primary domains.}
  \label{tab:appendix-benchmark-list}
\end{table}

\subsection{IRT fitting details}
\label{app:irt-fitting-details}

The IRT model is fit on the observed binary entries of the agent-task response matrix. We place zero-centered Gaussian priors with learned global scales on each agent ability \(\theta_a\) and task difficulty \(\beta_t\), approximate the posterior for each scalar with a normal variational factor, and optimize the evidence lower bound with stochastic variational inference. The posterior means are then used as the operational ability and difficulty estimates. The fitted scores \(\theta_a-\beta_t\) achieve response AUC \(0.937\) on observed pairs, indicating that the fitted scale captures the response matrix well enough to define the difficulty target used in the paper.

\paragraph{Fitting rationale.}
As in matrix factorization \citep{koren2009matrix}, only observed response-matrix entries contribute to the fit. A direct joint maximum-likelihood (JML) fit would initialize all abilities and difficulties and minimize binary cross-entropy over observed trials. This point-estimate fit is fragile under imbalance. Parameters with few observations become noisy, and the loss sees only the error in \(\theta_a-\beta_t\). If \(\theta_a\) is poorly estimated, a mispredicted trial can still push the paired \(\beta_t\) in the opposite direction, transferring noise between ability and difficulty.

The natural correction is to avoid committing to a single latent value too early. In the standard ability-marginalization view, for fixed task difficulties, each agent contribution is obtained by averaging over plausible abilities. Let \(\Omega_a=\{t:(a,t)\in\Omega\}\) be the observed tasks for agent \(a\). Then
\[
  p(Y_{a,\Omega_a}\mid\boldsymbol{\beta})
  =
  \int
  p(\theta_a)
  \prod_{t\in\Omega_a}
  p(Y_{a,t}\mid\theta_a,\beta_t)
  \,d\theta_a .
\]
A grid approximation with \(K\) support points requires \(O(K|\Omega|)\) likelihood evaluations per iteration, about \(4.15\times10^6\) evaluations for \(K=10\) at our scale, before repeated iterations. The variational fit replaces this grid enumeration with learned Gaussian approximate posteriors. The posterior means provide point estimates, while the posterior scales keep an uncertainty estimate for each ability and difficulty.

\paragraph{Implementation details.}
The fit uses a 1PL variational IRT implementation with Trace\_ELBO optimized by ClippedAdam for 2{,}000 epochs, learning rate \(0.01\), gradient clipping at norm \(5.0\), and random seed \(42\). The saved state contains \(\theta\), \(\beta\), and their posterior scales for 502 agents and 7{,}266 tasks before filtering to the main analysis subset.

\begin{figure}[h]
  \centering
  \includegraphics[width=0.72\linewidth]{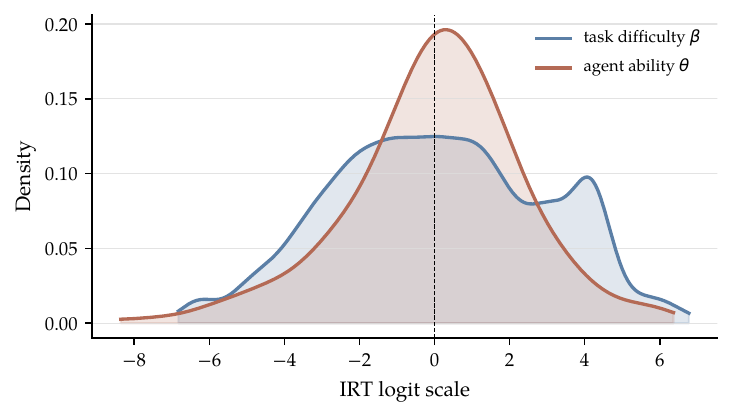}
  \caption{\textbf{Fitted IRT ability and difficulty distributions.} Kernel-density estimates are shown for item difficulty $\beta$ and agent ability $\theta$ before filtering to the main analysis subset.}
  \label{fig:irt-parameter-distribution}
\end{figure}

\subsection{Entropy scorer panel}
\label{app:entropy-scorers}

The single-scorer \emph{entropy} setting uses Qwen3-8B-Base. The \emph{full} setting uses the following five-scorer panel, in this fixed order for the scorer-count ablation: Qwen3-8B-Base; Qwen3-8B-Instruct, abbreviated as Qwen3-8B in Figure~\ref{fig:entropy-ablations} \citep{yang2025qwen3}; Gemma 4 E4B, abbreviated as Gemma-E4B-it in Figure~\ref{fig:entropy-ablations}; OLMo 3 7B Instruct, abbreviated as OLMo-3-7B-Inst \citep{olmo2025olmo}; and Llama 3.1 8B Instruct, abbreviated as Llama-3.1-8B-Inst \citep{grattafiori2024llama}.

\subsection{Estimator ablation}
\label{app:estimator-ablation}

Figure~\ref{fig:difficulty-prediction} fixes the prediction protocol to compare feature families. Table~\ref{tab:appendix-estimator-ablation} instead fixes the full feature representation and varies the regressor. Random forests improve KF performance slightly, reaching \(\rho=0.475\) and pairwise accuracy \(0.683\). Linear regression gives the strongest LOBO result, with \(\rho=0.295\) and pairwise accuracy \(0.602\). The ablation therefore leaves the main conclusion unchanged. Estimator choice affects interpolation within the pooled dataset, while benchmark transfer remains the limiting case.

\begin{table*}[t]
  \centering
  \small
  \setlength{\tabcolsep}{0pt}
  \begin{tabular*}{\textwidth}{@{\extracolsep{\fill}}lrrrrrr@{}}
    \toprule
    \textbf{Estimator} & \textbf{$N$} & \textbf{Dim.} & \textbf{KF $\rho \uparrow$} & \textbf{KF pAcc $\uparrow$} & \textbf{LOBO $\rho \uparrow$} & \textbf{LOBO pAcc $\uparrow$} \\
    \midrule
    Lin.  & 5{,}230 & 217 & $0.451 \pm 0.034$ & $0.676 \pm 0.012$ & $\boldsymbol{0.295 \pm 0.230}$ & $\boldsymbol{0.602 \pm 0.085}$ \\
    Ridge & 5{,}230 & 217 & $0.454 \pm 0.034$ & $0.678 \pm 0.013$ & $0.277 \pm 0.259$ & $0.597 \pm 0.101$ \\
    RF    & 5{,}230 & 217 & $\boldsymbol{0.475 \pm 0.033}$ & $\boldsymbol{0.683 \pm 0.015}$ & $0.130 \pm 0.250$ & $0.545 \pm 0.089$ \\
    XGB   & 5{,}230 & 217 & $0.453 \pm 0.037$ & $0.678 \pm 0.013$ & $0.190 \pm 0.199$ & $0.563 \pm 0.071$ \\
    \bottomrule
  \end{tabular*}
  \caption{Estimator choice does not remove the held-out-benchmark gap. Lin.=linear regression, RF=random forest, XGB=XGBoost. KF uses 10 folds; LOBO leaves out one benchmark. pAcc denotes within-benchmark pairwise accuracy. Entries report mean $\pm$ standard deviation across folds or held-out benchmarks.}
  \label{tab:appendix-estimator-ablation}
\end{table*}

\subsection{Human-time calibration}
\label{app:human-time-calibration}

Table~\ref{tab:appendix-human-time} calibrates fitted IRT difficulty to \(\log\) median human completion time on the subset with human-time annotations, following the BRIDGE framing \citep{liu2026bridge}. Linear and ridge regression perform nearly identically, with KF \(\rho=0.422\) and MAE \(0.857\). XGBoost gives the lowest KF MAE, \(0.849\), but weaker LOBO rank correlation. These results are auxiliary to the main paper, but they indicate that the fitted difficulty scale can be connected to human-interpretable task time when such annotations are available. Practically, accurate pre-rollout difficulty prediction could help researchers estimate human-time requirements more efficiently.

\begin{table*}[t]
  \centering
  \small
  \setlength{\tabcolsep}{0pt}
  \begin{tabular*}{\textwidth}{@{\extracolsep{\fill}}lrrrrr@{}}
    \toprule
    \textbf{Estimator} & \textbf{$N$} & \textbf{KF $\rho \uparrow$} & \textbf{KF MAE $\downarrow$} & \textbf{LOBO $\rho \uparrow$} & \textbf{LOBO MAE $\downarrow$} \\
    \midrule
    Lin.  & 1{,}676 & $\boldsymbol{0.422 \pm 0.060}$ & $0.857 \pm 0.016$ & $\boldsymbol{0.464 \pm 0.277}$ & $\mathbf{0.863}$ \\
    Ridge & 1{,}676 & $\boldsymbol{0.422 \pm 0.060}$ & $0.857 \pm 0.015$ & $\boldsymbol{0.464 \pm 0.277}$ & $\mathbf{0.863}$ \\
    RF    & 1{,}676 & $0.304 \pm 0.055$ & $0.922 \pm 0.026$ & $0.292 \pm 0.182$ & 0.957 \\
    XGB   & 1{,}676 & $0.408 \pm 0.056$ & $\boldsymbol{0.849 \pm 0.015}$ & $0.391 \pm 0.178$ & 0.886 \\
    \bottomrule
  \end{tabular*}
  \caption{Human-time calibration of fitted IRT difficulty, following the BRIDGE framing \citep{liu2026bridge}. Lin.=linear regression, RF=random forest, XGB=XGBoost. KF uses 5 folds. LOBO leaves out one benchmark. Uncertainties are standard deviations across folds. MAE is measured on the $\log$ median-human-minutes target.}
  \label{tab:appendix-human-time}
\end{table*}

\subsection{OLMo surprisal}
\label{app:olmo-surprise}

To sanity-check the use of language-model likelihood as a lightweight surprise signal, we scored a small set of prompt categories with \texttt{allenai/OLMo-1B-hf}. The experiment uses only short text snippets and should be read as a qualitative diagnostic rather than a benchmark result. For each prompt, we compute mean token surprisal, \(-\log p\), in nats per token under the model. Higher bars, therefore, indicate text that is more surprising to the scorer model.

\begin{figure}[h]
  \centering
  \includegraphics[width=0.78\linewidth]{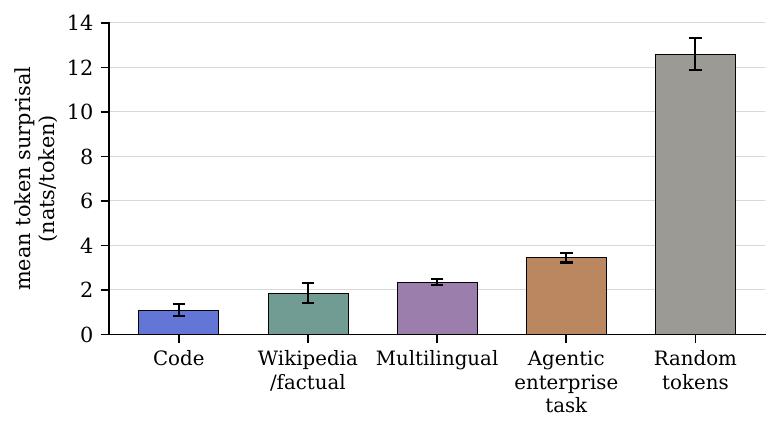}
  \caption{Mean token surprisal under OLMo-1B across prompt categories, measured in nats per token. Bars show means over five prompts per category; error bars show standard deviations. Random-token strings are included as a high-surprise control.}
  \label{fig:appendix-olmo-surprise}
\end{figure}

\end{document}